\documentclass{article}

 \usepackage[final]{neurips_2025_ml4ps}
 \makeatletter
\providecommand{\workshoptitle}[1]{\gdef\@workshoptitle{#1}}
\makeatother
\workshoptitle{Machine Learning and the Physical Sciences}

\usepackage[utf8]{inputenc} 
\usepackage{xcolor} 
\usepackage{graphicx} 
\usepackage[T1]{fontenc}    
\usepackage{hyperref}       
\usepackage{url}            
\usepackage{booktabs}       
\usepackage{amsfonts}       
\usepackage{nicefrac}       
\usepackage{microtype}      
\usepackage{spverbatim}
\usepackage{amsmath}

\title{Encoding and Understanding Astrophysical Information in Large Language Model-Generated Summaries}

\author{%
  Kiera McCormick \\
  Department of Computer Science\\
  Johns Hopkins University\\
  Baltimore, MD \\
  \texttt{kmccor23@jh.edu} \\
   \And
   Rafael Martínez-Galarza \\
   AstroAI \\
   Center for Astrophysics | Harvard \& Smithsonian \\
  \texttt{jmartine@cfa.harvard.edu} \\
}

\begin{document}

\maketitle

\begin{abstract}
Large Language Models have demonstrated the ability to generalize well at many levels across domains, modalities, and even shown in-context learning capabilities. This enables research questions regarding how they can be used to encode physical information that is usually only available from scientific measurements, and loosely encoded in textual descriptions. Using astrophysics as a test bed, we investigate if LLM embeddings can codify physical summary statistics that are obtained from scientific measurements through two main questions: 1) Does prompting play a role on how those quantities are codified by the LLM? and 2) What aspects of language are most important in encoding the physics represented by the measurement? We investigate this using sparse autoencoders that extract interpretable features from the text. 
\end{abstract}

\section{Introduction}

The current rise of Large Language Models (LLMs) has led researchers to begin incorporating them into higher scientific fields, through question-answer chat bots \cite{Iyer2024} \cite{Gao2024}, scientific document summarization and literature review \cite{Agarwal2024}, hypothesis generation \cite{Ciuca2023}, and more. One open question that relates to generalization in LLMs and alignment of text embeddings with other modalities is to what extent the representation space of LLMs encodes physical information that is obtained directly from scientific measurements and stored in numerical data structures. One example comes from astrophysics: observations, such as spectral measurements of astrophysical sources, encode information about the physical properties producing the electromagnetic emission. When humans describe those observations with language, are they successfully capturing the physics encoded in the data, and can LLMs capture that encoding?

Sparse Autoencoders (SAEs) have been used to find features that work as monosemantic units of analysis compared to the autoencoder neurons themselves, allowing for much more effective interpretability \cite{Bricken2023}. They have been used in the context of LLMs to investigate what specific tokens activate which features, creating a direct connection between words, concepts, and learned features. We build on this ability to investigate how specific concepts from text summaries correlate with summary statistics extracted from data objects of specific astrophysical objects. Specifically, we input text summaries describing astrophysical sources to pre-trained SAEs, and investigate how activating tokens relate to summary statistics obtained from the corresponding data objects.

In this work, we aim at understanding if and how LLMs can encode physics using astrophysical x-ray sources as a testbed. We investigate two main questions: (i) To what extent does prompt engineering alone affect the level of correlation between text embeddings and physical quantities? and (ii) How exactly is the information about physics encoded in text summaries? More specifically, what are the concepts that drive the correlation with physical properties extracted from data? All code is open-sourced and can be found \href{https://github.com/kieramccormick/Encoding-Physical-Info}{here}.

\section{Related works}
In 2023, tools for prompt engineering, few-shot learning, and embedding tasks for astronomy were released \cite{Nguyen2023AstroLLaMA} \cite{Sotnikov2023}. Wu et al. 2024 designed a Retrieval-Augmented Generation powered chat bot for research astronomy \cite{Wu2024} \cite{Iyer2024}. Anthropic released papers that attempted to understand how LLMs work using SAEs, including their Golden Gate Claude \cite{Templeton2024} \cite{Bricken2023}. Research into disentangling dense embeddings with sparse autoencoders was released in 2024 \cite{O'Neill2024}. 

\section{Data preparation}
This section describes how the dataset used in this system was created and how to reproduce this experiment within other scientific fields.

\begin{figure}
    \centering
    \includegraphics[width=1\linewidth]{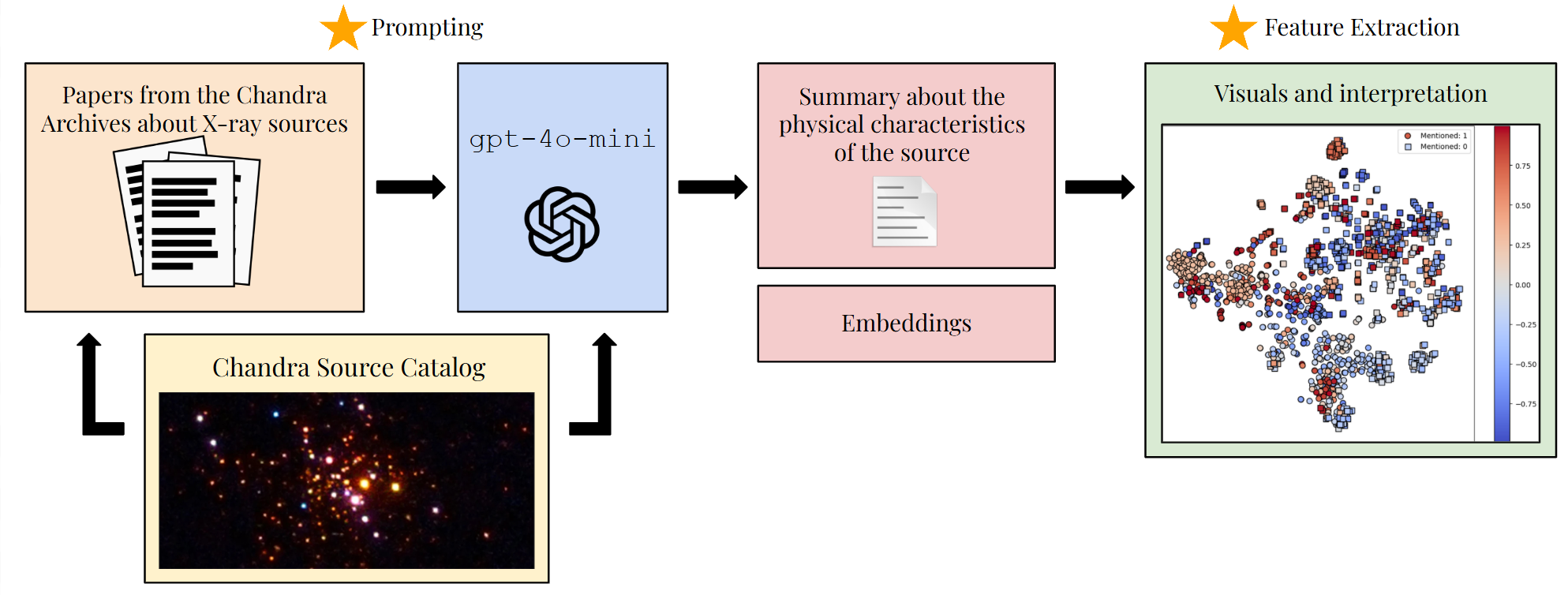}
    \caption{This figure gives an overview of how this dataset was created. Starting with a Chandra observation, we queried the Chandra Archives for papers about each specific source within the observation, or each bright area within the yellow image. Then we fed these papers to \texttt{gpt-4o-mini} and asked it to create a summary about the physical characteristics of the source. The code also generated embeddings of these summaries and several of the physical parameters of these objects. Finally, we used these embeddings and summaries to create visuals of the physical information for interpretation. The two stars seen in this diagram emphasize what this work focuses on: prompt engineering and feature extraction.}
    \label{fig:methods}
\end{figure}

We use two sources of data: a text corpus curated by the NASA Astrophysics Data System (ADS) \cite{ads} consisting of scientific papers that include descriptions of specific X-ray observations of astrophysical sources, and physical summaries associated with these sources extracted from the Chandra Source Catalog (CSC)\footnote{https://cxc.cfa.harvard.edu/csc/} (Figure \ref{fig:methods}). We prompted OpenAI's \texttt{gpt-4o-mini} model to search for physical information in the ADS papers associated with each specific source within a Chandra telescope observation using a combination of source sky coordinates and source identifiers associated with each unique object.  

The LLM creates a summary of the physical information about the source, and the CSC provides relevant physical summary statistics, including measurements that encode the X-ray spectral shape of the source. One example is the slope of the power-law model fitted to the X-ray spectrum. The LLM summaries are then encoded using the \texttt{ada-002} model. This results in a set of LLM embeddings for each source's textual summary, paired with physical quantities extracted from the source's astrophysical measurement. Two examples of output summaries can be found in Appendix \ref{app:summaries}.

The physical summary statistics include the hardness ratio, power law gamma value, variability index, and more that can be found in Appendix \ref{app:physicalqualities}. This metadata is used to understand and assign semantic meaning to the clustering of text embeddings. If there is a relationship between the language describing the source and the measured physical quantities, we would expect that clusters of text embeddings would also cluster in terms of their physical quantities.

\section{Encoding physical information} \label{sec:prompting}
The section of this work will discuss how to best encode physical information in LLM-generated summaries. 

\subsection{Creating a prompt}

We investigate whether the specific wording of the prompt affects the level of alignment between text embeddings and the physical properties. We compared two different prompts, which can be found in Appendix \ref{app:prompts}. The first prompt asked the LLM to use the context provided to summarize specific physical properties related to the spectrum and the time variability of the source \cite{Martinez-Galarza2025}. The second prompt asked for the same information in a more structured fashion with a block of formatting instructions, guidelines for how to handle missing information, and more explicit instructions about the specific information that the LLM is looking for in the context. The second prompt also attempts to avoid encoding text that does not provide actual physical information, such as “the Hardness Ratio of the source was not mentioned in this text”. The first prompt analyzed was used as a baseline for improvements, and the second prompt was the product of experimenting. 

\subsection{Understanding the effects of prompting}

\begin{figure}
    \centering
    \includegraphics[width=1.0\linewidth]{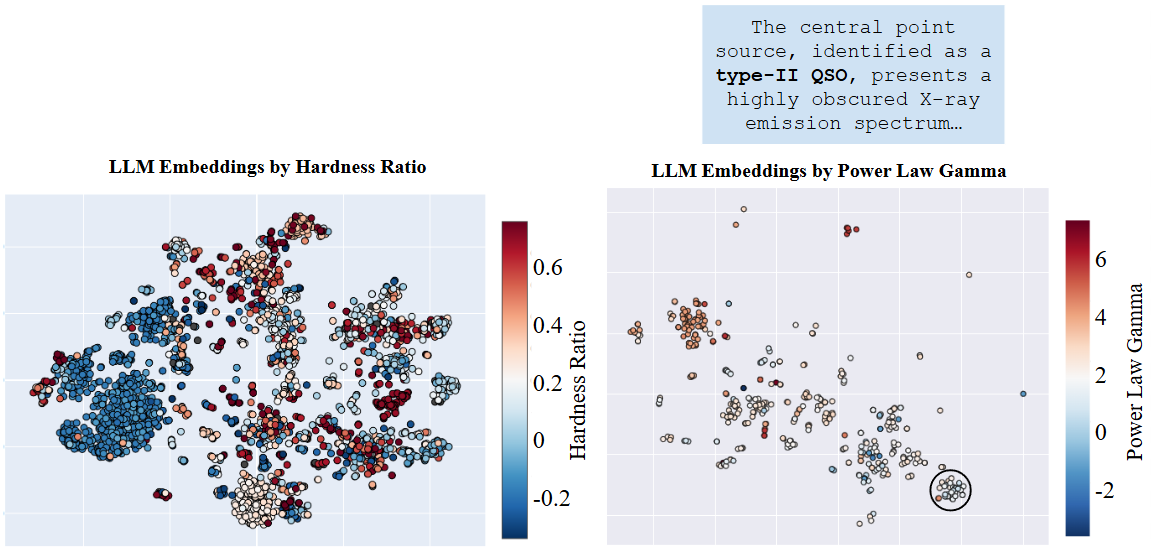}
    \caption{An example of visually understanding where the LLM is encoding physical information in the embeddings using t-SNE plots. The left plot is colored by hardness ratio, which is a broader summary statistic, with around 4000 samples. The right plot is in relation to power law gamma value with about 1700 samples, and the above text is an excerpt from a specific source summary within the circled cluster. Power law gamma is a specific spectral property, and while the explicit power law gamma value was never mentioned in the text summary, the LLM was able to infer and learn about this physical property from the language in the summary.}
    \label{fig:PLG}
\end{figure}

The two different prompts were compared in terms of the level of correlation between the resulting summary embeddings and the physical quantities extracted from the telescope data. In order to investigate this correlation, we performed dimensionality reduction using t-SNE on the original \texttt{ada-002} embeddings and assessed clustering quality using k-nearest neighbors purity analysis (k=10), which quantifies whether sources with similar physical properties are positioned near each other in the embedding space.  Figure \ref{fig:PLG} shows two examples of how the resulting LLM embeddings correlate with the slope of the hardness ratio or power law fit to the spectrum created from the experimental prompt. Visually, the resulting clusters in the LLM embedding space are more discriminative in terms of physical property when the second, more structured prompt is used. Quantitatively, the updated prompt demonstrated substantially improved clustering quality from the baseline, original prompt. K-nearest neighbors analysis revealed significant improvements in clustering purity: hardness ratio improved from 0.7998 to 0.8468 (5.9\%), power law gamma improved from 0.8185 to 0.9418 (15.1\%), and most notably, variability index improved from 0.6346 to 0.9994 (57.5\%). This showcases that having detailed and specific prompts works better for uncovering the learning ability of \texttt{gpt-4o-mini} for highly specialized physical phenomena, and results in a more informative LLM response.

Figure \ref{fig:PLG} demonstrates how to visually understand how the LLM embeddings encode information about physical measurements. The specific values of the physical parameters, such as the power law slope or hardness ratio, are not typically explicitly mentioned in the text. In the general case, the LLM does not have information about the numerical value of the physical parameters. Yet, the t-SNE clustering in Figure \ref{fig:PLG} indicates that the relevant information about the physical parameters is registered in the text, probably because the pre-trained LLM understands that certain types of astrophysical objects tend to display specific values of these parameters. Or because the language that describes the astrophysical properties is enough to infer the numerical values of measured quantities, implying that the LLM can infer some physics concepts. Other parameters tested included black body temperature and variability probability, which also displayed clustering purity improvements from the baseline to experimental prompt, indicating further that prompting does play an important role in correlating text embeddings and physical quantities. 

\section{Extracting physical information using sparse autoencoders}
In order to investigate how language encodes relevant information, as the correlation between summary text embeddings and physical measurement suggests, we use SAEs. In regular autoencoders, a single neuron can be activated by multiple different input features, which makes interpretability difficult. SAEs address this by restricting neuron activation so that each neuron is likely to represent a distinct and monosemantic feature, forcing the most meaningful aspects of the data to be captured.

We use pre-trained SAEs, a 1-layer \texttt{gelu-1l} language model trained to reconstruct the activations of a one-layer transformer with a 512-neuron MLP layer, from \cite{Bricken2023}. We perform inference with the pre-trained transformer by inputting the dataset of LLM-generated summaries. We then observe the hidden layer to see which of the learned features were more prominent in each of the LLM embeddings clusters of Figure \ref{fig:PLG}, and to what contexts those features were activated in. 

From Figure \ref{fig:PLG}, we selected several LLM embedding clusters whose members also shared a similar value of the measurement statistic, and investigated which SAE features activated those clusters to uncover what semantic concepts in the text were underlying to the cluster structure. Let us take the circled cluster in Figure \ref{fig:PLG} as an example. We chose the top five strongest and most unique features in each cluster to analyze. The features that the SAEs find are unlabeled, so we used both handmade note taking as well as prompting \texttt{Claude Sonnet 4} and \texttt{Google Gemini Pro 2.5} to label these features. A table with feature labels is included in Appendix \ref{app:features}, and an example of a feature with the top words that activate it is in Appendix \ref{app:SAEExample}. Our SAE visualization tool outputs the top twenty words that fire for each feature, as well as the snippet of the LLM-generated summary where that token appears for context.

We find that each cluster associated with a particular range of power-law slopes is also associated with some specific concepts, and that such concepts are relevant in codifying the measured physical quantities. For example, the majority of the Chandra sources in one cluster (highlighted in Appendix \ref{app:features}) are associated with Eclipsing Binaries, which are stars that orbit each other with such orientation that they eclipse each other periodically from the perspective of Earth. In the circled cluster in Figure \ref{fig:PLG}, many of the features related to non-thermal sources, which correlates with having a lower power law gamma value. The LLM was able to infer this information and encode having a lesser power law gamma value, even though it was not explicitly mentioned in the text summaries.

In Figure \ref{fig:PLG}, the bottom cluster seemed to all have sources with non-thermal characteristics. One specific source within this cluster was described as being a type-II QSO, and references "energetic outbursts from the AGN", which both support having non-thermal characteristics. So while the power law gamma value was never mentioned in the text, or whether the source was thermal or not, the model was still able to understand that since a source was a certain type of object, it would have a specific physical parameter value due to the other characteristics of the source. This proves that the model is learning and understanding complex astrophysical information, and even making inferences.

\section{Conclusions and future work}
LLMs have demonstrated their ability to capture, learn, and represent physics concepts in textual data. The textual and numerical data about the physical properties of astrophysical X-ray sources can be aligned and correlated through prompt engineering. Using SAEs to create monosemantic features allows researchers to investigate what in the text informs the LLM about the physical parameters of these sources and offers insight towards what physics concepts these models are learning.

In future works, this dataset will be expanded upon as well as prompting with different models for optimum encoding. An interface for astronomer validation of possibly anomalous objects is currently being created, in which the expert will be provided with a text-based summary and the accompanying spectra.  

\section*{Acknowledgments}

This work was conducted during the AstroAI summer internship program at the Center for Astrophysics | Harvard \& Smithsonian. KM thanks AstroAI for their support. 



\bibliographystyle{plain}   
\bibliography{references}


\appendix

\section{Appendix: Example Summary Outputs}\label{app:summaries}

This section displays two output summary examples: a well-informed one and a less-informed one. This is meant to give a better idea of how this system is working and how the LLM is encoding and "understanding" this physical information. The well-informed example is from the middle of the red cluster in Figure \ref{fig:plgplot} that is defined by stellar objects and variability, and the less-informed example is the paler point directly above this cluster. 

\textbf{Well-Informed Summary:}

A) X-ray Properties

The source in question exhibits a significant degree of variability, specifically identified as a flaring object with an observed peak flux density increase during periods of outburst. At its highest observed intensity, the flux density at 86 GHz increased more than fivefold within a matter of hours, reaching a peak of 160 mJy. This observed variability includes repeated occurrences over a span of roughly 70 days, although subsequent flares were of lesser intensity than the initial discovery. The characteristics demonstrate a decay pattern; after the initial outburst, flux levels decayed over days, with reported measurements going as low as approximately 11 mJy.

The spectral properties of the source were assessed through various observations including X-ray emission. The best-fit spectral model for the X-ray flux indicated an intrinsic X-ray luminosity \(L_x\) of about \(10^{31.7}\) erg s\(^{-1}\), which was attenuated by a column density of \(N_H = 10^{22.6}\) cm\(^{-2}\). This suggests the presence of a significant absorbing medium around the source, impacting its observed luminosity.
Timing analysis indicates variability on short timescales with peak emissions during flares, described as having rise times of approximately one hour. While the exact orbital period is not definitively stated, variations were correlated with the rotational phase of the source, allowing implications about underlying binary interactions or rotational dynamics.

Multi-wavelength data also indicate the source's position as coincident with optical and infrared measurements, suggesting it could be classified as a young stellar object within a more complex environment. The near-infrared spectrum revealed that the source is consistent with being a K5V star and showed signs of magnetic activity, corroborated by Zeeman splitting measurements that inferred a magnetic field of \(B \approx 2.6 \pm 1.0\) kG.

B) Use in Scientific Hypotheses

The properties of the source are utilized effectively to test several astrophysical hypotheses, particularly the mechanisms of stellar magnetic activity in young stellar objects. The demonstrated flaring behavior supports models of active magnetic fields associated with such stars, leading to further understanding of the physics governing stellar rotation, magnetism, and accretion processes.

The significant increase in X-ray luminosity and the associated spectral changes during flares contribute to confirming theories about coronal heating and dynamics in young, active stellar environments. The relationship between X-ray emissions and radio flaring strengthens the association between magnetic fields and the acceleration of particles, as it was observed that peaks in X-ray count rates occurred shortly before radio flares, thereby constraining models of magnetic reconnection processes in young stellar objects.

Overall, these observations collectively advance the understanding of the evolution and behavior of stars in stellar nurseries, as well as the fundamental processes driving star formation dynamics and magnetic influences in accreting systems.

\textbf{Less-Informed Summary:}

A) X-ray Properties

The source exhibits X-ray properties characteristic of early-type stars (type Er*), which are often associated with strong magnetic activity and high X-ray luminosities. Although specific data about variability such as transient behavior or periodicity is not provided in the text, it is implied that such stars may exhibit typical behaviors like flares or variations in quiescent states as seen in analogous early-type stars.

Spectral properties of the source can be described with observational data suggesting that X-ray emission is often modeled with parameters such as a power-law spectrum. Specific best-fit parameters related to similar sources typically include a power-law photon index (\( \Gamma \)), disk temperature (\( kT_{\text{in}} \)), and column density (\( N_H \)); however, no numerical values or concrete uncertainties are given in this document.

Flux measurements commonly recorded for such sources fall within the ranges typical for young, active stars. The total luminosity is usually significantly high due to the impact of magnetic activities. Multi-wavelength data, such as optical magnitudes or infrared measurements, are not explicitly mentioned but can generally provide critical supporting evidence in X-ray studies for such objects.

B) Use in Scientific Hypotheses

Properties of the source are essential in context to testing or constraining scientific models related to the formation and evolution of planetary systems, particularly in understanding how X-ray-driven photoevaporation affects protoplanetary discs. Enhanced X-ray emissions from a stellar source can influence the thermal and dynamical states of surrounding material, which is crucial during the initial phase of planetary formation.

This information may be instrumental to accretion processes, specifically in determining the mechanisms that drive the mass loss from disks, shaping the environment for planet formation and influencing the final architecture of planetary systems. Moreover, insights into the magnetic activity of such stars could reveal correlations with the presence of potential planetary companions, shedding light on the dynamics of star-planet interactions and their implications for planetary habitability.

In summary, while specific quantitative measurements for this source are not elaborately listed in the text, information surrounding X-ray characteristics and their implications point towards a significant role in the development of young stellar systems and their associated planets.

\section{Appendix: Astrophysical Properties}\label{app:physicalqualities}

Table \ref{tab:astrophysicalqualities} is a list of astrophysical properties extracted from the Chandra Source Catalog and appended in the created embeddings along with the LLM-generated summaries. Definitions are from the CSC.

 \begin{table}[htb]
     \centering
     \begin{tabular}{p{4cm}|p{10cm}}
     \toprule
     \textbf{Parameter} & \textbf{CSC Definition} \\
    \midrule
\texttt{hard\_hs} & ACIS hard (2.0-7.0 keV) - soft (0.5-1.2 keV) energy band hardness ratio \\
 \midrule
 \texttt{bb\_kt} & temperature (kT) of the best fitting absorbed black body model spectrum to the source region aperture PI spectrum \\
 \midrule
 \texttt{powlaw\_gamma} & photon index, defined as $F_E \propto E^{-\gamma}$, of the best fitting absorbed power-law model spectrum to the source region aperture PI spectrum \\
 \midrule
 \texttt{var\_index\_b} & index in the range [0,10] that combines (a) the Gregory-Loredo variability probability with (b) the fractions of the multi-resolution light curve output by the Gregory-Loredo analysis that are within $3\sigma$ and $5\sigma$ of the average count rate, to evaluate whether the source region flux is uniform throughout the observation \\
 \midrule
 \texttt{prob\_index\_b} & N/A \\
 \bottomrule

    \end{tabular}
    \caption{Astrophysical properties that were included in the embeddings and metadata files created by this system. For each source, these values were recorded to be used for understanding how they were recorded in the generated summaries.}
    \label{tab:astrophysicalqualities}
\end{table}

\section{Appendix: Prompts}\label{app:prompts}

Below are the two prompts used in this paper.

\textbf{Original Prompt:}
\begin{spverbatim}
    Given the text provided, search for information about the source 
    identified with any of the following names:
    {', '.join(repr(item) for item in name_ids)}. 

    The source is a source of type {tipos[j]}.

    Again based on the text provided, answer the following questions 
    regarding the source in question, without mentioning the name of the 
    source or the target:

    Is the source specifically mentioned in the text, or is the source the 
    target of the observation? If the answer is 'yes' to any of these 
    questions, do the following. 
    If not, do the following in reference to sources of type {tipos[j]} 
    described in the text.
        A) Summarize the X-ray properties of the source in question, as 
        inferred directly from the data. Focus on variability (transient 
        behavior, periodicity, etc.), and spectral features (models fitted, 
        hardness ratios, n_h, etc.), but provide values of any relevant 
        measured quantities if measured directly from the X-ray data.
        B) Describe how these properties or other X-ray data from the 
        source is used to test the scientific hypotheses being examined in 
        the text provided.
\end{spverbatim}

\textbf{Updated Prompt:}
\begin{spverbatim}
    Within the text you are provided with, search for information about the 
    source identified with any of the following names:
    {', '.join(repr(item) for item in name_ids)}.
    The source is classified as type {tipos[j]}.
    
    Your task is to extract and summarize the physical properties and 
    scientific interpretation of this source as completely and directly as 
    possible, using only information contained in the text.

    Please first evaluate whether the source is directly mentioned in the 
    text. If the source is mentioned directly (or is the target of the 
    observation), return:  
    [MENTIONED: YES]  
    Otherwise return:  
    [MENTIONED: NO] 
    Then, provide the full physical summary as before.
    If 
    the source is not directly mentioned or targeted, provide a general 
    summary based on the information available for sources of type 
    {tipos[j]}.
    
    Follow these instructions strictly:
    ### A) X-ray Properties
    - Describe variability, including:
        - Transient behavior, periodicity, flares, quiescence, outbursts
        - Decay patterns (exponential decay, e-folding times, linear decay 
        rates)
        - Orbital periods (report estimates if available)
    - Spectral properties:
        - Spectral models fitted (e.g., power-law, disk blackbody, 
        Comptonization)
        - Best-fit parameters (e.g., photon index Gamma, disk temperature 
        kT_in, column density N_H)
        - Include all provided uncertainties with numerical values.
        - Report state transitions (e.g., hard state, thermally dominated, 
        steep power law)
        - Hardness ratios (if provided)
    - Flux measurements and luminosity (always include units where 
    possible)
    - Timing analysis (variability timescales, periodicities, orbital 
    periods)
    - Multi-wavelength data (e.g., optical magnitudes, IR, radio 
    measurements if stated)
    - Include any specific values explicitly reported in the text.
    
    ### B) Use in Scientific Hypotheses
    - Describe how these properties are used to test or constrain 
    scientific models discussed in the text.
    - Include discussion of accretion processes, black hole or neutron star 
    identification, coronal structure, super-Eddington behavior, binary 
    evolution, or any astrophysical interpretation directly stated.
    
    ### Formatting Instructions:
    - Present your answer in complete sentences, fully written out, but 
    remain clear and concise.
    - Always prioritize quantitative measurements when available.
    - Include all physical properties mentioned, even if multiple models or 
    parameters are provided.
    - Do not speculate beyond the information provided.
    - Do not refer to the source by name or target name.
\end{spverbatim}

\section{Appendix: SAE feature labels}\label{app:features}

\begin{figure}
   \centering
   \includegraphics[width=1\linewidth]{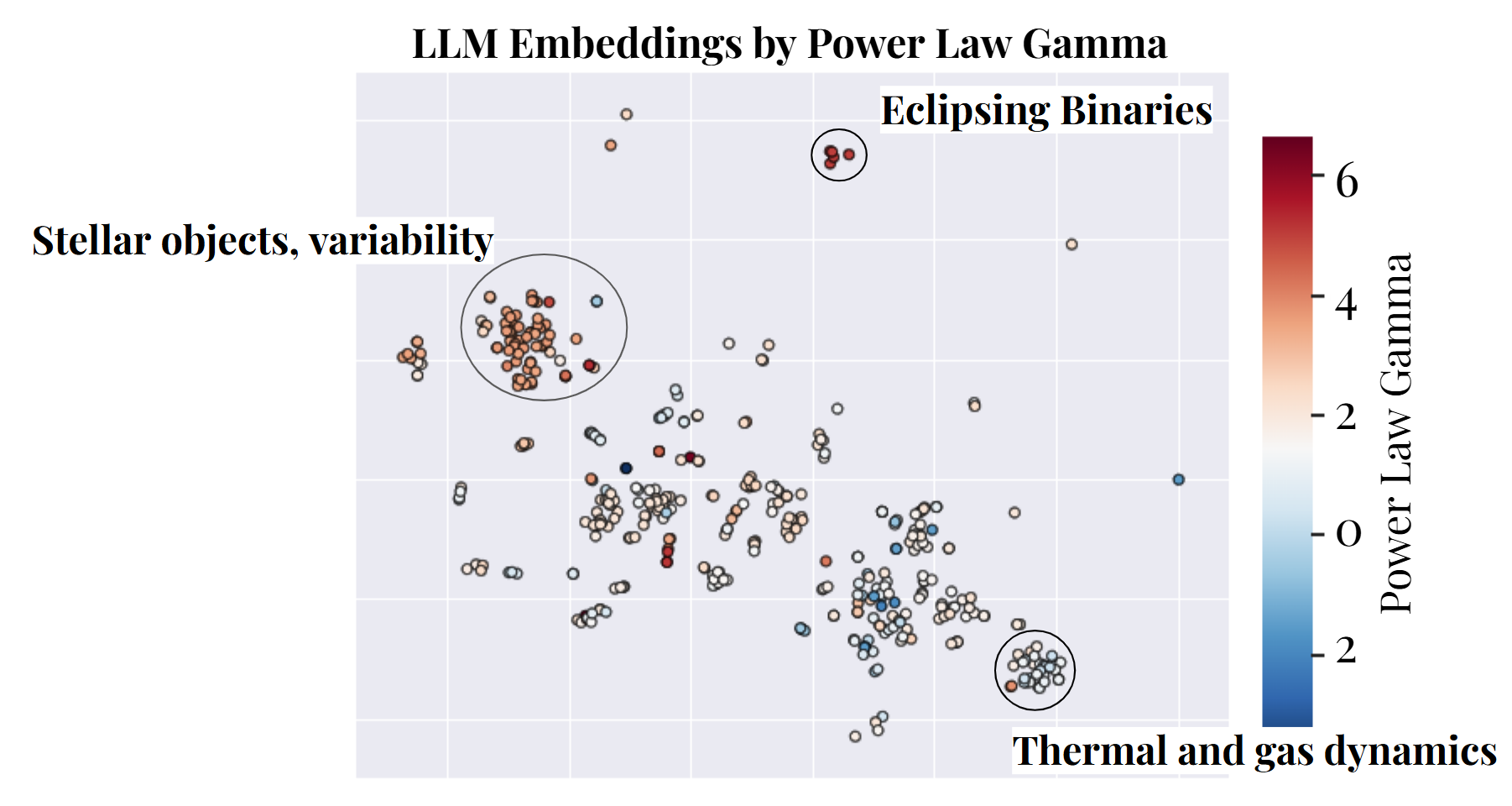}
   \caption{t-SNE plot of source summaries colored by power law gamma. Each of the three isolated clusters was analyzed using SAEs to understand the underlying themes within the summaries that allowed them to cluster. The text by each summary highlights the themes found across the features.}
   \label{fig:plgplot}
\end{figure}

Table \ref{tab:clusterlabels} shows the top two unique and strongest features for each of the three clusters. The three clusters can be seen in Figure \ref{fig:plgplot}.

\begin{table}[htb]
    \centering
    \begin{tabular}{p{1cm}|p{4.5cm}|p{4.5cm}|p{4.5cm}}
    \toprule
    \textbf{Feature} & \textbf{Cluster 1} & \textbf{Cluster 2} & \textbf{Cluster 3} \\
   \midrule
1212 & Scientific Inference and Possibility & & \\
\midrule
839 & X-ray Binary Outburst Cycles & & \\
\midrule
411 & & Galaxy Cluster Thermal Dynamics and Cooling Processes & \\
\midrule
1262 & & X-ray Astrophysical Source Characteristics & \\
\midrule
1181 & & & X-ray Source Variability and Emission Properties \\
\midrule
1160 & & & Methodological Analysis of Variable Astrophysical Sources \\
\bottomrule

   \end{tabular}
   \caption{The top two strongest and unique features across the three clusters seen in Figure \ref{fig:plgplot}.}
   \label{tab:clusterlabels}
\end{table}

\section{Appendix: SAE Feature Example}\label{app:SAEExample}

 \begin{figure}
     \centering    
     \includegraphics[width=1\linewidth]{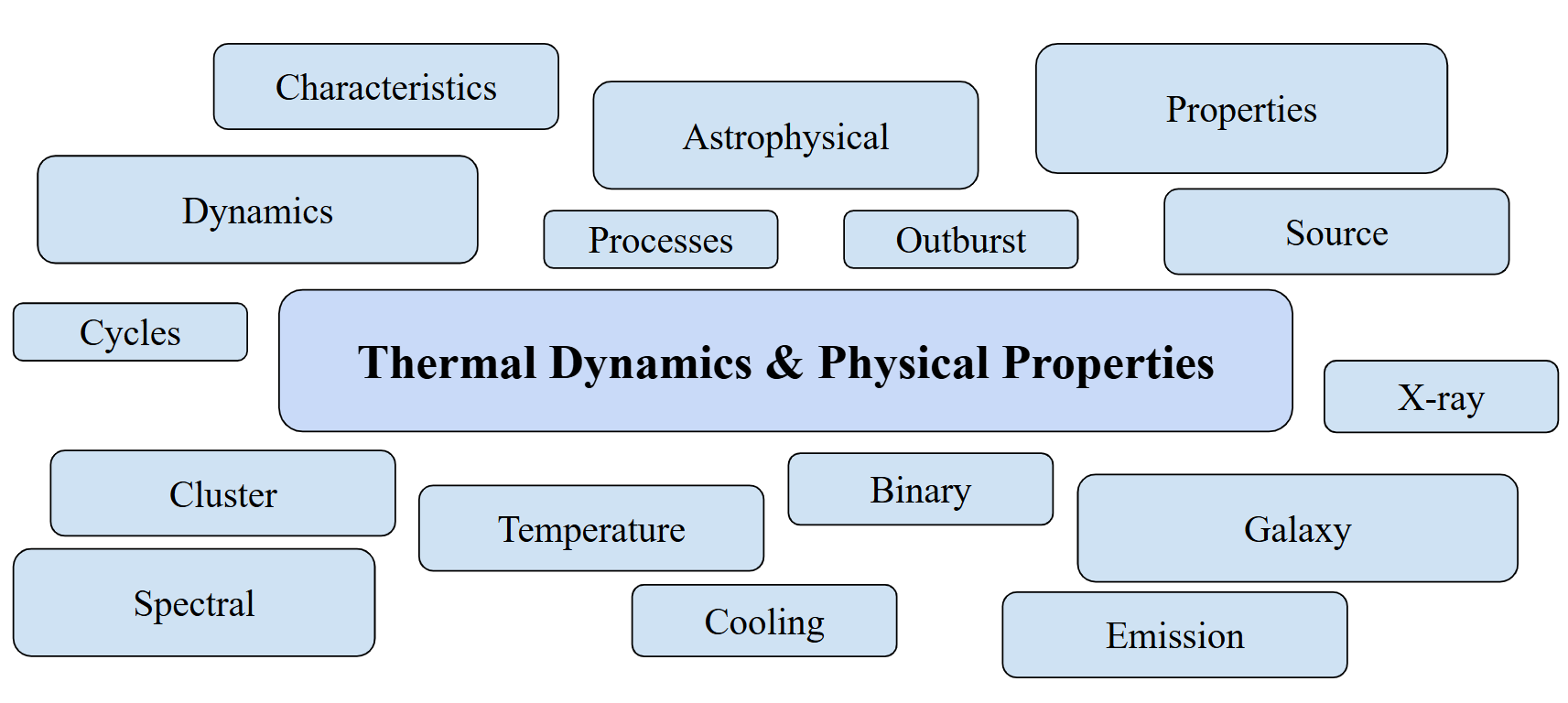}
     \caption{A label found by SAEs with the top words that activate for it. A similar feature was found across many clusters, and the semantic details within each feature changed for each differing dataset used as context.}
     \label{fig:CloudCluster}
 \end{figure}

 Figure \ref{fig:CloudCluster} showcases a feature found in several clusters with the top words that activate for it. Each cluster examined a unique dataset of astrophysical sources and their generated summaries, and SAEs found a similar top feature in each. However, this similar feature relates to "Thermal Dynamics", and these specific dynamics differed for each cluster. For example, the highlighted cluster in Figure \ref{fig:PLG} had Feature 411, labeled "Galaxy Cluster Thermal Dynamics and Cooling Processes", which discussed non-thermal dynamics that related to the sources in this cluster, while a similar feature in other clusters talked about thermal dynamics.

\end{document}